\documentclass[11pt]{article}

\usepackage{xeCJK} 
\usepackage{latexsym}
\usepackage{amsmath,amssymb,amsthm}
\usepackage{graphicx}
\usepackage{booktabs}
\usepackage{multirow}
\usepackage{url}
\usepackage{hyperref}
\usepackage{tikz}
\usetikzlibrary{arrows.meta,positioning,fit,shapes.geometric}
\usepackage{algorithm}
\usepackage{algorithmic}

\usepackage[margin=1in]{geometry}

\newtheorem{observation}{Observation}
\newtheorem{remark}{Remark}

\title{From Association to Causation: Improving Retrieval Precision of\\
Retrieval-Augmented Generation via Causal Relations and an Attention Mechanism}

\author{Jing Liu, Yongxing Qi, Muchen Jiang, Chengnan Hu, Qingqing Peng, \\Haoming Wang, Yuqing Wang, Yang Yu, Xu Zhang, Ting Wu\\
Hangzhou Innovation Institute, Beihang University, Hangzhou, China\\
\texttt{hiiliujing@buaa.edu.cn}}

\date{}

\begin{document}
\maketitle

\begin{abstract}
Retrieval-Augmented Generation (RAG) grounds LLM generation on retrieved documents, but the standard terminal retrieval stage---dense-vector similarity, optionally followed by reranking---often returns documents that merely share keywords with the query without containing the needed information, a failure mode that grows with the knowledge base. We trace it to a conceptual gap: similarity captures only \emph{associational} relations, whereas the documents that matter are linked to the query \emph{causally}. We model the terminal retrieval stage with a causal graph grounded in Reichenbach's common cause principle: the keywords shared by the query and a retrieved document form a latent common cause $A$, and the document's residual keywords form a latent set $B$ linking the document to the ideal output. Since a retrieved document is a collider ($A \!\rightarrow\! d \!\leftarrow\! B$), retrieval itself opens an associational path between the query and $B$, which licenses a training-free, attention-style re-scoring rule: the cosine similarity between the query embedding and the weighted centroid embedding of $B$. Unlike causality-enhanced RAG variants that model causal relations \emph{inside} the knowledge content at the cost of LLM-built graphs and extra LLM calls, our graph models the causal structure of the \emph{retrieval process itself}. On a real 471-document enterprise knowledge base, the method promotes a genuinely relevant guideline from rank 6 to the top 3; on a controlled diagnostic corpus reproducing the keyword-stuffing regime, it improves the mean target rank from 2.88 to 1.25, while a trained cross-encoder reranker barely helps (2.63). Conversely, on three BEIR benchmarks the score underperforms the similarity baseline, delineating the applicability boundary: the method guards the keyword-stuffing regime of growing proprietary knowledge bases and complements neural rerankers; a corpus-level calibration gate selects the correct regime with $\ge 95\%$ reliability. A fully local Qwen3-4B/BGE-M3 testbed demonstrates practical deployability.
\end{abstract}

\section{Introduction}

Large language models (LLMs) such as the GPT family have advanced rapidly and attracted worldwide attention. When deployed in vertical domains, however, LLMs suffer from hallucination, and enterprise data is often too sensitive to be used directly for fine-tuning. Retrieval-Augmented Generation (RAG)~\cite{lewis2020retrieval,gao2023ragsurvey} has therefore become the standard architecture for knowledge-intensive applications: internal documents and knowledge bases are first vectorized; given a user query, relevant documents are retrieved and injected into the prompt of the LLM, improving the accuracy and reliability of the generated content.

A conventional RAG system consists of (i) a \emph{retriever} that returns a set of candidate text fragments from a large corpus, typically via vector similarity or semantic matching; (ii) a \emph{generator}, usually a Transformer-based LLM, that produces a coherent answer conditioned on the query and the retrieved fragments; and (iii) a \emph{fusion mechanism}---concatenation, attention, or related schemes---that combines the retrieved fragments with the generator input.

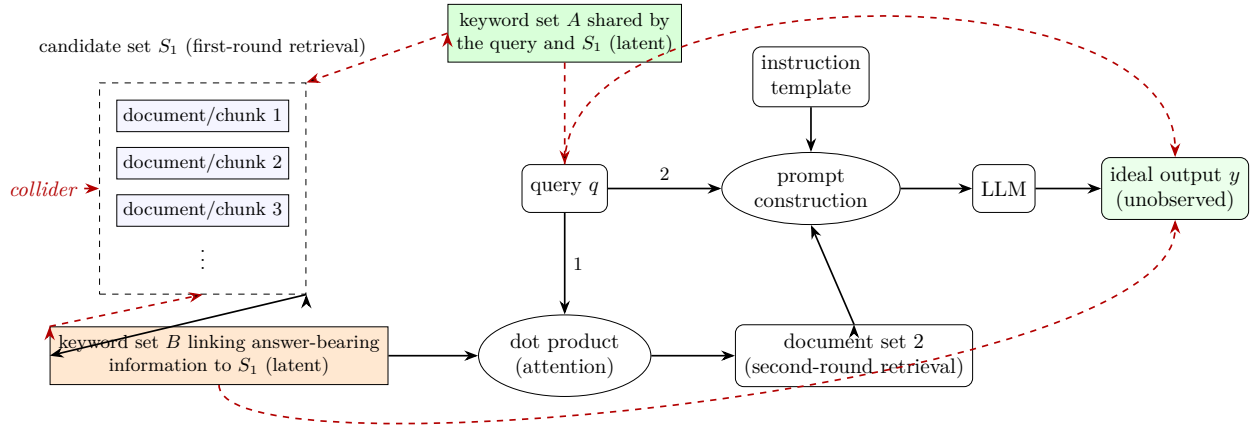
\begin{figure}[t]
\centering
\resizebox{\textwidth}{!}{%
\begin{tikzpicture}[
  flow/.style={rectangle, draw, rounded corners, align=center, font=\small, inner sep=4pt, minimum height=0.8cm},
  doc/.style={rectangle, draw, align=center, font=\footnotesize, inner sep=3pt, fill=blue!4},
  latentA/.style={rectangle, draw, align=center, font=\footnotesize, inner sep=4pt, fill=green!15},
  latentB/.style={rectangle, draw, align=center, font=\footnotesize, inner sep=4pt, fill=orange!18},
  circ/.style={ellipse, draw, align=center, font=\small, inner sep=3pt},
  dataflow/.style={draw, -{Stealth}, thick},
  causal/.style={draw=red!70!black, dashed, -{Stealth}, thick},
  node distance=0.8cm and 1.1cm]
\node[doc] (d1) {document/chunk 1};
\node[doc, below=0.25cm of d1] (d2) {document/chunk 2};
\node[doc, below=0.25cm of d2] (d3) {document/chunk 3};
\node[font=\footnotesize, below=0.05cm of d3] (ddots) {$\vdots$};
\node[draw, dashed, inner sep=8pt, fit=(d1)(d2)(d3)(ddots)] (s1) {};
\node[font=\footnotesize, above=0.3cm of s1] (s1lab) {candidate set $S_1$ (first-round retrieval)};
\node[font=\small\itshape, text=red!70!black, left=0.25cm of s1] (collab) {collider};
\draw[causal] (collab.east) -- (s1.west);
\node[flow, right=3.6cm of s1] (q) {query $q$};
\node[latentA, above=1.7cm of q] (A) {keyword set $A$ shared by\\ the query and $S_1$ (latent)};
\node[circ, below=1.7cm of q] (dot) {dot product\\ (attention)};
\node[latentB, left=1.5cm of dot] (B) {keyword set $B$ linking answer-bearing\\ information to $S_1$ (latent)};
\node[flow, right=1.4cm of dot] (s2) {document set 2\\ (second-round retrieval)};
\node[circ, right=1.9cm of q] (prompt) {prompt\\ construction};
\node[flow, above=0.75cm of prompt] (tmpl) {instruction\\ template};
\node[flow, right=1.2cm of prompt] (llm) {LLM};
\node[flow, right=1.1cm of llm, fill=green!8] (y) {ideal output $y$\\ (unobserved)};
\draw[dataflow] (q) edge node[right, font=\footnotesize]{1} (dot);
\draw[dataflow] (s1.south east) edge (B.west);
\draw[dataflow] (B) edge (dot);
\draw[dataflow] (dot) edge (s2);
\draw[dataflow] (s2.north) edge (prompt.south);
\draw[dataflow] (q) edge node[above, font=\footnotesize]{2} (prompt);
\draw[dataflow] (tmpl) edge (prompt);
\draw[dataflow] (prompt) edge (llm);
\draw[dataflow] (llm) edge (y);
\draw[causal] (A) edge (q);
\draw[causal] (A.west) edge (s1.north east);
\draw[causal] (B.north west) edge (s1.south);
\draw[causal] (q.north) .. controls +(0,3.2) and +(0,3.2) .. (y.north);
\draw[causal] (B.south) .. controls +(0,-1.6) and +(0,-2.8) .. (y.south);
\end{tikzpicture}}
\caption{Unified view of the proposed method at the terminal retrieval stage of RAG (redrawn in English from the flow--causal diagram in the underlying invention disclosure~\cite{patent}). Solid arrows: data flow. Red dashed arrows: causal directions among the query $q$, the first-round candidate set $S_1$, the ideal output $y$, and the two latent keyword sets $A$ and $B$ (cf.\ the abstract causal graph in Figure~\ref{fig:causal}). Step 1 scores each candidate by the attention-style dot product between the query embedding and the centroid of its residual keyword set $B$; step 2 constructs the prompt from the re-ranked set. The candidate set $S_1$ is a collider ($A \!\rightarrow\! d \!\leftarrow\! B$), which justifies the score of step 1 (Observation~\ref{obs:collider}).}
\label{fig:flow}
\end{figure}

The weak point of this architecture is the end of the retrieval stage (Figure~\ref{fig:flow}). Both industry and academia predominantly compute the similarity between the query vector and knowledge-base vectors, optionally followed by reranking. Through empirical study we observed that, as the knowledge base keeps growing, this pipeline increasingly surfaces documents that contain the query \emph{keywords} but are not truly relevant, so the genuinely useful local information never reaches the LLM. For example, for the query ``How to assess the security of cross-border data transfer?'' (数据出境的安全如何评估), a conventional retriever ranked an interpretation of an automotive data-processing guideline within the top 5, simply because it contains more occurrences of the keywords ``data'', ``cross-border'', ``security'', and ``assessment''. The genuinely relevant national guideline \emph{Information Security Technology---Guidelines for Security Assessment of Cross-Border Data Transfer} was ranked only 6th: it mentions these keywords mainly in section titles, while the body paragraphs that actually describe the assessment procedure use different vocabulary.

We trace this failure to a conceptual gap: similarity-based retrieval identifies \emph{association}, whereas what the user needs is the document that \emph{causally} contains the answer. This paper makes the following contributions:

\begin{itemize}
\item We formalize the terminal retrieval stage of RAG with a causal graph built on Reichenbach's common cause principle, and prove a collider-opening observation: retrieving a document ($A \!\rightarrow\! d \!\leftarrow\! B$) induces an association between the query and the document's residual keyword set $B$---the set that carries the answer-bearing vocabulary (Section~\ref{sec:causal}).
\item We derive from the graph a practical, training-free re-scoring rule with the computational form of attention---the cosine similarity between the query embedding and the weighted centroid embedding of latent keyword set $B$---together with a complexity analysis showing that the overhead over a conventional pipeline is negligible (Section~\ref{sec:method}).
\item We position the method against recent work on adaptive, corrective, graph-structured, and causality-enhanced RAG (Section~\ref{sec:related}), showing that it is, to the best of our knowledge, the only approach that models the causal structure of the retrieval process itself rather than of the knowledge content.
\item We show that the construction is consistent with the standard assumptions of causal graphical models (DAG, causal sufficiency, causal Markov, and faithfulness), and evaluate the method on three levels: a real 471-document proprietary knowledge base, a controlled diagnostic experiment reproducing the keyword-stuffing regime, and three public BEIR benchmarks that delineate its applicability boundary, together with a fully local reproduction testbed based on Qwen3-4B and BGE-M3 (Section~\ref{sec:exp}).
\end{itemize}

An early version of the method was disclosed in a Chinese invention patent application filed in 2024, since granted~\cite{patent}. The present paper gives the full formalization, the theoretical analysis, and the empirical evaluation, including the semantic-absorption refinement of Section~\ref{sec:method} and the public-benchmark boundary study of Section~\ref{sec:beir}, which go beyond the disclosure.

\section{Related Work}
\label{sec:related}

\paragraph{RAG and its retrieval stage.} RAG was popularized by \cite{lewis2020retrieval} and pretraining-time variants such as REALM~\cite{guu2020realm} and RETRO~\cite{borgeaud2022retro}, and has become the standard grounding mechanism for LLMs~\cite{gao2023ragsurvey}. Work on the retrieval stage improves the encoder (DPR~\cite{karpukhin2020dense}, Contriever~\cite{izacard2021unsupervised}, BGE and BGE-M3~\cite{bge2023,bgem3}) and the reranker (cross-encoders, ColBERT-style late interaction~\cite{khattab2020colbert}). All of these optimize an associational score estimated from correlated features; none models why a document is relevant.

\paragraph{Adaptive and corrective RAG.} A recent line of work lets the system decide \emph{when} and \emph{what} to retrieve and how to recover from retrieval errors: Self-RAG teaches the generator to retrieve on demand and self-critique via reflection tokens~\cite{asai2024selfrag}; FLARE performs forward-looking active retrieval during generation~\cite{jiang2023flare}; CRAG grades retrieved documents and routes low-confidence results to corrective actions~\cite{yan2024crag}; RA-DIT dual-instruction-tunes the retriever and generator~\cite{lin2024radit}. These methods improve robustness around the retrieval stage but still score candidates by learned association; the criterion that decides what enters the prompt remains correlational.

\paragraph{Query expansion and hypothetical documents.} A classical remedy for vocabulary mismatch is to expand the query: relevance feedback and Rocchio's method~\cite{rocchio1971relevance}, relevance models (RM3)~\cite{lavrenko2001relevance}, and, more recently, LLM-generated pseudo-content such as HyDE~\cite{gao2023hyde} and Query2doc~\cite{wang2023query2doc}. These methods augment the \emph{query} with hypothetical answer-bearing content and embed the result. Our construction is the dual viewpoint: rather than hallucinating content on the query side, we distill each candidate \emph{document} into a residual keyword set whose involvement is justified by a causal graph, and score it against the unmodified query. The two directions are complementary and could be combined.

\paragraph{Term-centric and learned sparse retrieval.} Our use of keyword sets is related to term-centric neural IR: DeepCT re-weights terms by contextual importance~\cite{dai2019deepct}, doc2query/DocT5Query expand documents with predicted queries~\cite{nogueira2019doc2query}, and the SPLADE family learns sparse term-weight representations end to end~\cite{formal2021splade}. These methods learn term weights from relevance supervision, i.e., from associational signals; we instead \emph{construct} the term sets ($A$, $B$) from the causal structure of the retrieval process, which is training-free and works on top of any encoder.

\paragraph{Graph-structured and causality-enhanced RAG.} GraphRAG builds an LLM-extracted entity graph with community summaries for query-focused global sensemaking~\cite{edge2024graphrag}; LightRAG combines graph indexing with dual-level retrieval for efficiency~\cite{guo2024lightrag}; HippoRAG builds a knowledge-graph memory with Personalized PageRank for associative retrieval~\cite{gutierrez2024hipporag}. Closest to our theme, CausalRAG integrates causal graphs into retrieval by matching the query to graph nodes, expanding along edges, and using an LLM to trace causal paths \emph{within the document content} into a causal summary~\cite{wang2025causalrag}; Samarajeewa et al.\ similarly retrieve causal graphs in a pre-retrieval stage~\cite{samarajeewa2024causal}. The key difference is where the causality lives: these methods model cause--effect relations \emph{inside the knowledge content} (e.g., \emph{influence tactics} $\rightarrow$ \emph{buyer attention} $\rightarrow$ \emph{contract award}), at the price of LLM-built graphs and additional LLM calls at query time. We model the causal structure of the \emph{retrieval process itself}---the generative relations among the query, the retrieved set, and the ideal output---and derive a scoring rule from a collider analysis of that structure. The two are complementary: our re-ranker could post-process the candidate sets produced by any of these systems.

\paragraph{Causal inference.} Reichenbach's common cause principle states that if two events are positively correlated and neither causes the other, there exists a common cause that screens them off~\cite{reichenbach}. Modern causal discovery formalizes such reasoning through causal graphs under the causal Markov and faithfulness assumptions~\cite{spirtes2000causation,pearl2009causality,spirtes2016causal,shimizu2006linear}, and recent work probes how far LLMs can reason from correlation to causation~\cite{jin2024corr2cause}. A collider ($X \rightarrow Z \leftarrow Y$) blocks the path between its parents marginally, but conditioning on it induces dependence between them---the ``explaining away'' effect~\cite{pearl2009causality}. In recommender systems, causal structure has been used to \emph{remove} bias, e.g., intervening on popularity confounders via backdoor adjustment~\cite{zhang2021causal}. We use causal structure in the opposite direction and with no adjustment at all: we exploit the dependence that a collider \emph{creates} between the query and the residual keyword set.

\paragraph{Positioning.} Taken together, existing work improves RAG retrieval along associational axes---better encoders, rerankers, query expansion, adaptivity, or graph structure---while causality-enhanced variants model causality inside the knowledge content and pay for it with LLM-built graphs and per-query LLM calls. To the best of our knowledge, no prior work models the causal structure of the retrieval process itself or derives a re-scoring rule from a collider/common-cause analysis of that structure. The present method is simultaneously causally motivated, training-free, free of extra LLM calls at query time, and orthogonal to all of the above; its target failure mode---keyword-stuffing false positives in a growing knowledge base---is the regime where purely associational scores degrade.

\section{Method}

\subsection{Problem Setting and Notation}

Let $q$ be a user query and $\mathcal{K}$ a knowledge base of $N$ documents. A standard terminal retrieval stage computes dense embeddings $E(\cdot) \in \mathbb{R}^{h}$ (e.g., $h{=}1{,}024$ for BGE-M3), ranks documents by cosine similarity\begin{equation}
s_{\mathrm{sim}}(d) \;=\; \cos\!\big(E(q),\, E(d)\big),
\end{equation}
optionally applies a reranker, and returns the candidate set
\begin{equation}
S_1 \;=\; \mathrm{top}\text{-}k(\mathcal{K},\, s_{\mathrm{sim}}) \;\cap\; \big\{ d \in \mathcal{K} : s_{\mathrm{sim}}(d) \ge \theta \big\},
\end{equation}
where $\mathrm{top}\text{-}k(\mathcal{K}, s_{\mathrm{sim}})$ returns the $k$ documents with the highest scores and $\theta$ is a similarity threshold (document set~1). Throughout, $E(s)$ denotes the encoder's embedding of any text span $s$; in particular, for a keyword term $t$, $E(t)$ is the embedding of that term treated as a short text. The failure mode we address: $S_1$ contains both the documents that genuinely hold the needed information and documents that merely share surface keywords with $q$. Our goal is to re-rank $S_1$ so that genuinely informative documents are promoted. We write $\mathrm{kw}(\cdot)$ for a weighted keyword extractor that maps a text to a set of (term, weight) pairs $\{(t, \lambda_t)\}$; the instantiation is flexible (TF--IDF, TextRank, KeyBERT, or an LLM-based extractor) and is not the contribution of this work. Duplicate terms returned by the extractor (e.g., surfacing from different chunks) are collapsed into a single entry with their weights summed, so $K_q$ and $K_d$ below are proper sets.

\subsection{A Causal View of Terminal Retrieval}
\label{sec:causal}

We introduce two latent variables defined over keyword sets:

\begin{itemize}
\item \textbf{Latent set $A$} --- the keywords shared by the query $q$ and a retrieved document $d \in S_1$. Both $q$ and $d$ can be viewed as generated from $A$ (they are \emph{about} these keywords), so $A$ is a common cause: $A \rightarrow q$ and $A \rightarrow d$.
\item \textbf{Latent set $B$} --- the document keywords that remain after removing $A$, together with their weights. These keywords connect the retrieved document to the information the user actually wants; $B$ generates both the document and the ideal output text $y$: $B \rightarrow d$ and $B \rightarrow y$.
\end{itemize}

\begin{figure}[t]
\centering
\begin{tikzpicture}[
  latent/.style={circle, draw, minimum size=1.15cm, inner sep=1pt},
  obs/.style={rectangle, draw, minimum size=1.0cm, inner sep=3pt},
  every edge/.style={draw, -{Stealth}, thick}]
\node[latent] (A) {$A$};
\node[obs, below left=1.5cm and 2.2cm of A] (q) {$q$};
\node[obs, below right=1.5cm and 2.2cm of A] (d) {$d \in S_1$};
\node[latent, right=2.8cm of A] (B) {$B$};
\node[obs, below right=1.5cm and 2.2cm of B] (y) {$y$};
\draw (A) edge (q);
\draw (A) edge (d);
\draw (B) edge (d);
\draw (B) edge (y);
\draw (q) edge[bend right=25] (y);
\end{tikzpicture}
\caption{Causal graph of the terminal retrieval stage. $q$: user query; $d$: a document of the initially retrieved set $S_1$; $y$: the ideal output text (unobserved); $A$, $B$: latent keyword sets. $d$ ($A \!\rightarrow\! d \!\leftarrow\! B$) and $y$ ($q \!\rightarrow\! y \!\leftarrow\! B$) are colliders. Marginally $q \perp B$; conditioning on the retrieved document $d$ opens the path $q \!\leftarrow\! A \!\rightarrow\! d \!\leftarrow\! B$ (Observation~\ref{obs:collider}), licensing the attention-style score of Section~\ref{sec:method}.}
\label{fig:causal}
\end{figure}
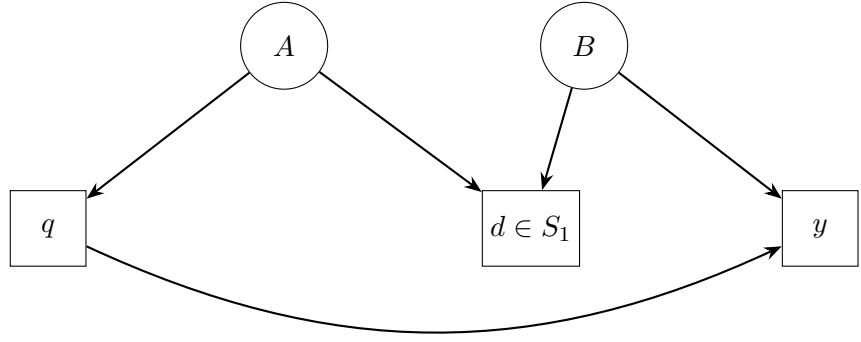

The \emph{ideal output} $y$ is the answer an ideal system would produce; it is unobserved at retrieval time. Since $y$ is generated from the query and keyword set $B$, we have $q \rightarrow y \leftarrow B$; and since a retrieved document is generated from both latent keyword sets, we have $A \rightarrow d \leftarrow B$. Both $d$ and $y$ are therefore \emph{colliders} in the graph (Figure~\ref{fig:causal}).

Two structural facts follow. First, \emph{screening off}: conditioning on the common cause $A$ renders $q$ and $d$ independent (the remaining path $q \!\rightarrow\! y \!\leftarrow\! B \!\rightarrow\! d$ is blocked by the unconditioned collider $y$), and conditioning on both latent variables $\{A, B\}$ renders $d$ and $y$ independent (the paths $d \!\leftarrow\! B \!\rightarrow\! y$ and $d \!\leftarrow\! A \!\rightarrow\! q \!\rightarrow\! y$ are blocked at the conditioned forks). This matches Reichenbach's characterization of common causes. Second, and operationally more important, is the collider-opening effect:

\begin{observation}[Retrieval opens the query--$B$ path]
\label{obs:collider}
In the graph of Figure~\ref{fig:causal}, under the causal Markov and faithfulness assumptions:
\begin{enumerate}
\item[(i)] $q$ and $B$ are marginally independent: the only two paths between them, $q \leftarrow A \rightarrow d \leftarrow B$ and $q \rightarrow y \leftarrow B$, are both blocked by colliders ($d$ and $y$ respectively).
\item[(ii)] $q$ and $B$ are dependent given $d \in S_1$: conditioning on the retrieved document (the collider) opens the path $q \leftarrow A \rightarrow d \leftarrow B$.
\end{enumerate}
\end{observation}

\begin{proof}[Proof sketch]
Enumerate the $q$--$B$ paths in Figure~\ref{fig:causal}. Marginally, $d$ is a collider on $q \!\leftarrow\! A \!\rightarrow\! d \!\leftarrow\! B$ and $y$ is a collider on $q \!\rightarrow\! y \!\leftarrow\! B$; both paths are blocked, so $q \perp B$ by d-separation, and independence follows by the Markov assumption. Given $d$, the collider on the first path no longer blocks it, and $A$ is an unconditioned fork on that path; hence the path is active, $q \not\perp B \mid d$ by d-connection, and dependence follows by faithfulness.
\end{proof}

Observation~\ref{obs:collider} is the central inference based on the causal analysis: \emph{once a document has been retrieved, the query and the document's residual keyword set $B$ are expected to be associated}, and this association is informative because $B$ excludes the keywords $A$ that merely make the document look superficially similar to the query. Retrieval itself acts as the conditioning event that turns this association on.

\begin{remark}[Scope of the causal argument]
\label{rem:scope}
Our use of the causal graph is structural and motivational rather than inferential: we do not estimate causal effects from data, and the score $s_c$ defined below is not an identified causal estimand. The graph provides (i) a principled reason, via Observation~\ref{obs:collider}, to expect $q$ and $B$ to be associated over answer-bearing documents, and (ii) a construction rule for the residual keyword set. Selecting documents into $S_1$ by a score computed from $d$ is an idealized form of conditioning on $d$; we read the result as a heuristic justification for the scoring rule, with the empirical burden carried by Section~\ref{sec:exp}.
\end{remark}

\subsection{Consistency with Causal-Model Assumptions}
\label{sec:assumptions}

The construction is consistent with the standard assumptions of causal graphical models, which lends it a principled interpretation rather than an ad-hoc one:

\begin{enumerate}
\item \textbf{Directed acyclic graph (DAG) assumption.} The relations among the research objects ($q$, $d$, $y$, $A$, $B$) are described by a DAG (Figure~\ref{fig:causal}); the generative direction flows from latent keyword sets to observed texts, with no feedback cycles.
\item \textbf{Causal sufficiency assumption.} All direct common causes of any two variables in the variable set are included in the set~\cite{spirtes2016causal,druzdzel2009role}: the two latent keyword sets $A$ and $B$ are explicitly modeled as the common causes, and remaining influences are treated as independent exogenous variables.
\item \textbf{Causal Markov assumption.} Given a causally sufficient variable set, every variable is independent of its non-descendants conditional on its parents~\cite{spirtes2000causation}; the DAG in Figure~\ref{fig:causal} is a causal graph iff the joint distribution of its nodes satisfies this Markov condition, which the generative story (texts generated from keyword sets) respects by construction.
\item \textbf{Causal faithfulness assumption.} Conditional independencies in the joint distribution correspond exactly to d-separations in the graph~\cite{pearl2009causality,shimizu2006linear}; no accidental cancellations are assumed, so the collider-induced association of Observation~\ref{obs:collider} is a genuine structural property rather than an artifact of a parameterization.
\end{enumerate}

\subsection{Scoring Rule and Algorithm}
\label{sec:method}

For each candidate document $d \in S_1$, let $K_q = \mathrm{kw}(q)$ and $K_d = \mathrm{kw}(d)$ be the weighted keyword sets. The latent sets are instantiated as
\begin{equation}
A(d) \;=\; \big\{ t \in K_d : t \in K_q \ \lor\ \max_{u \in K_q} \cos\!\big(E(t), E(u)\big) \ge \tau \big\}, \qquad B(d) \;=\; K_d \setminus A(d),
\end{equation}
i.e., $A$ absorbs not only exact keyword matches but also document keywords that are \emph{semantically} near the query keywords (absorption threshold $\tau$; we use $\tau{=}0.6$ in the experiments), and $B$ is the document's \emph{residual} vocabulary once the shared query--document keywords have been absorbed into $A$. The semantic-absorption matters in practice: exact string matching leaves near-synonymous distractor vocabulary in $B$, which measurably degrades the score (Section~\ref{sec:diagnostic}). We embed $B$ into a single vector by its weighted centroid,
\begin{equation}
v_B(d) \;=\; \frac{\sum_{(t,\lambda_t) \in B(d)} \lambda_t\, E(t)}{\sum_{(t,\lambda_t) \in B(d)} \lambda_t},
\end{equation}
i.e., the weighted mean of the term embeddings---the standard way to aggregate word vectors with weights~\cite{arora2017sif}. The disclosure underlying this work specifies pooling over the keyword embeddings without fixing the pooling form; the weighted centroid is our instantiation, using the keyword weights that the extraction step provides.
and define the \emph{causal-attention score} as
\begin{equation}
s_c(d) \;=\; \cos\!\big(E(q),\, v_B(d)\big).
\end{equation}
By Observation~\ref{obs:collider}, $s_c$ measures exactly the association that the causal graph predicts to exist between the query and the answer-bearing residual keywords. It is an attention mechanism whose query--key pairing is causally motivated rather than heuristic. Documents that merely repeat the query's keywords have small or empty $B$ (everything informative was absorbed into $A$) and are demoted; documents whose residual vocabulary carries the answer receive high $s_c$ and are promoted. Algorithm~\ref{alg:ours} summarizes the full refinement stage, which appends to a conventional pipeline (Figure~\ref{fig:flow}).

\begin{algorithm}[t]
\caption{Causal-attention re-ranking for RAG terminal retrieval}
\label{alg:ours}
\begin{algorithmic}[1]
\REQUIRE query $q$; knowledge base $\mathcal{K}$; embedding model $E(\cdot)$; keyword extractor $\mathrm{kw}(\cdot)$; LLM
\STATE Compute $E(q)$ and $E(d)$ for $d \in \mathcal{K}$ and form the candidate set $S_1$ by Eqs.~(1)--(2) (threshold $\theta$, optional reranker).
\STATE Extract the query keyword set $K_q \leftarrow \mathrm{kw}(q)$ (keyword set 1).
\FOR{each document (or chunk) $d \in S_1$}
  \STATE Extract the document keyword set $K_d \leftarrow \mathrm{kw}(d)$ with weights (keyword set 2).
  \STATE Construct $A(d)$ and $B(d)$ by Eq.~(3); \quad record $|A|$.
  \STATE Compute $v_B(d)$ and $s_c(d)$ by Eqs.~(4)--(5).
\ENDFOR
\STATE Re-rank $S_1$ by $s_c$ (optionally interpolated with $s_{\mathrm{sim}}$ or $|A|$; a design choice, cf.\ Section~\ref{sec:discussion}); output the top documents as the genuinely informative set.
\STATE Build the prompt from the selected documents with the instruction template and feed it to the LLM.
\end{algorithmic}
\end{algorithm}

\paragraph{Complexity.} The refinement adds, per query, $k$ keyword extractions and $k \cdot \bar{b}$ term embeddings, where $k = |S_1|$ (single digits to tens in practice) and $\bar{b}$ is the average size of the residual keyword sets (tens of terms). This is negligible compared with the $O(N)$ document embeddings of the initial retrieval over a corpus of $N \gg k$ documents, requires no training or fine-tuning, and---in contrast to graph- or causality-enhanced RAG variants---adds no LLM calls at query time.

\subsection{Implementation}

A reference implementation was built in Python: the base LLM and the BGE-M3 embedding model are loaded (S1); instruction templates, document loading, data cleaning, foreign-language translation, document-splitting, and keyword-extraction submodules are implemented (S2); the local knowledge base is vectorized with BGE-M3 (S3); the candidate set $S_1$ is retrieved and re-ranked by the causal-attention score (S4--S8); and the final prompt is constructed and passed to the LLM (S9). The pipeline is model-agnostic: the original deployment used a 6B-parameter open-source chat model (ChatGLM3-6B), and we additionally rebuilt the full infrastructure on a laptop with Qwen3-4B as the base model (which also serves as the local translation module for foreign-language documents, replacing an earlier commercial translation API) and BGE-M3 for embeddings (Section~\ref{sec:testbed}).

\section{Experiments}
\label{sec:exp}

\subsection{Setup and Evaluation Protocol}

The knowledge base contains 471 local files (2.95\,GB), spanning laws and regulations, internal reports, professional books, and academic papers, mostly Word and PDF documents. Documents are parsed, cleaned (removal of table-of-contents dot leaders, URLs, and parsing artifacts), split into chunks of 4{,}096 characters with 512-character overlap, and embedded with BGE-M3 (1{,}024-dimensional dense vectors, mean-pooled per file). The baseline terminal retrieval computes query--document cosine similarity, applies a similarity threshold, and takes the top-$k$ documents---the standard industrial practice. Our method re-ranks this candidate set as in Algorithm~\ref{alg:ours}.

We report the retrieval outcome as the \emph{rank position of the a-priori-known target document}---the document that domain inspection identifies as actually containing the answer---under the baseline and under the proposed re-ranking. The evaluation is a real-deployment case study on a proprietary corpus; quantitative evaluations follow in Sections~\ref{sec:diagnostic}--\ref{sec:gating}.

\subsection{Case Study: Cross-Border Data Transfer Security}
\label{sec:case}

Consider the real user query ``How to assess the security of cross-border data transfer?'' (如何评估数据出境的安全). Table~\ref{tab:case} shows the baseline top-8 and our top-3 on the original deployment. We note for precision that the 2024 deployment used the exact-match $A/B$ construction (keyword-set intersection as in the invention disclosure~\cite{patent}); the semantic-absorption refinement of Eq.~3 was developed later, motivated by the failure analysis in Section~\ref{sec:diagnostic}, and is the form we recommend.

\begin{table}[t]
\centering
\small
\begin{tabular}{cl}
\toprule
Rank & Document \\
\midrule
\multicolumn{2}{l}{\emph{Baseline (similarity + rerank), top-8}}\\
1 & Measures for Security Assessment of Cross-Border Data Transfer (draft)\\
2 & Measures for Security Assessment of Personal Information and Important Data\\
  & Cross-Border Transfer (draft)\\
3 & Measures for Security Assessment of Personal Information Cross-Border Transfer (draft)\\
4 & Interpretation of \emph{Technical Guidelines for Automotive Collected-Data}\\
  & \emph{Processing Security} \\
5 & Technical Guidelines for Automotive Collected-Data Processing Security\\
6 & \textbf{Information Security Technology---Guidelines for Security Assessment}\\
  & \textbf{of Cross-Border Data Transfer} \\
7 & Several Provisions on Automotive Data Security Management (trial)\\
8 & Data Security Law of the P.R.C.\\
\midrule
\multicolumn{2}{l}{\emph{Ours (causal-attention re-ranking), top-3}}\\
1 & Measures for Security Assessment of Personal Information and Important Data\\
  & Cross-Border Transfer (draft)\\
2 & Measures for Security Assessment of Cross-Border Data Transfer (draft)\\
3 & \textbf{Information Security Technology---Guidelines for Security Assessment}\\
  & \textbf{of Cross-Border Data Transfer} \\
\bottomrule
\end{tabular}
\caption{Retrieval results for the query ``How to assess the security of cross-border data transfer?''. Bold: the document that actually contains the assessment procedure (《信息安全技术 数据出境安全评估指南》). The baseline buries it at rank 6, behind keyword-matching automotive-data documents; our method ranks it within the top 3. Document titles translated from Chinese; the draft measures at ranks 1--3 (baseline) and 1--2 (ours) are also relevant regulations, while ranks 4, 5, and 7 of the baseline concern a different domain (automotive data).}
\label{tab:case}
\end{table}

The baseline's failure is diagnostic. The automotive-data documents mention ``data'', ``cross-border'', ``security'', and ``assessment'' repeatedly in their bodies, so their dense vectors are close to the query vector; the genuinely relevant guideline instead concentrates these terms in section titles and describes the actual assessment procedure with different vocabulary. Similarity ranking therefore rewards keyword density, not answer-bearing content. Under the proposed method, the shared set $A$ (``data'', ``cross-border'', ``security'', \dots) absorbs exactly the keywords that made the automotive documents look relevant, and the residual set $B$ exposes the difference: the automotive documents' $B$ contains domain-specific terminology unrelated to assessment procedures, while the guideline's $B$ carries the assessment-procedure vocabulary. The causal-attention score $s_c$ consequently promotes the guideline from rank 6 into the top 3, and demotes the three automotive documents out of the returned set entirely.

\subsection{Reproduction Testbed}
\label{sec:testbed}

To make the pipeline inspectable and reproducible without any proprietary service, we re-implemented the full infrastructure to run entirely on a local machine: Qwen3-4B (4B parameters) serves both as the generator and as the local translation module for foreign-language documents, and BGE-M3 provides embeddings. The testbed performs PDF/Word parsing, cleaning, chunking, thresholded similarity retrieval with single-document and cross-document chunk strategies, on-the-fly translation of non-Chinese retrieved chunks, and prompt-based generation. The causal re-ranker of Algorithm~\ref{alg:ours} is implemented on top of the same embedding backbone and is evaluated in Section~\ref{sec:diagnostic}. The complete testbed source code, together with all experiment and analysis scripts reported in this section, is publicly available at \url{https://github.com/Silk-Road/causal-rag-rerank}.

Two sanity checks confirm the testbed's end-to-end behavior. First, on a mixed-language corpus of books and technical documents, the Chinese query ``according to traditional Chinese medicine, how should persistent cough be treated, and what medicine should be taken?'' correctly retrieved the formulary volume \emph{A Practical Handbook of TCM Formulas} (《实用趣味方剂手册》) from the corpus, and the local model produced a syndrome-differentiated answer grounded in the retrieved chunks. Second, the local translation module correctly rendered English retrieval chunks into Chinese with terminology annotations (e.g., translating a passage on the Wason selection task from a psychology-of-reasoning volume). These checks confirm that the infrastructure is model-agnostic: it exhibits the same end-to-end behavior with the original ChatGLM3-6B deployment and with the Qwen3-4B testbed.

\subsection{Controlled Diagnostic Experiment}
\label{sec:diagnostic}

To make the keyword-stuffing failure reproducible and inspectable under known ground truth, we built a small controlled corpus of short documents spanning eight themes (five Chinese, three English), each containing one a-priori-known \emph{target} document that carries the answer, near-relevant documents, deliberately keyword-stuffed distractors (documents that repeat the query keywords without answering the query), and unrelated documents: (T1) cross-border data-transfer security assessment, mirroring the case study of Section~\ref{sec:case}; (T2) TCM treatment of cough; (T3) findings of the Wason selection task; (T4) LLM pretraining-data cleaning; (T5) personal-information protection impact assessment; (T6) ColBERT's late interaction; (T7) TCM treatment of insomnia; (T8) SPLADE's regularization. The corpus is synthetic by construction and serves to demonstrate the mechanism, not to estimate effect sizes. The keyword extractor $\mathrm{kw}(\cdot)$ is instantiated with jieba segmentation + TF--IDF weights for Chinese and a word tokenizer + TF--IDF for English (top-15 terms per document, top-10 per query); $\tau{=}0.6$ for semantic absorption. We compare five rankings of the candidate set: the similarity baseline $s_{\mathrm{sim}}$; a strong trained cross-encoder reranker (BGE-reranker-v2-m3~\cite{bgem3}) applied to the same candidates; the causal-attention score $s_c$ with \emph{exact-match} $A/B$ construction; $s_c$ with \emph{semantic absorption}; and a hybrid $\tfrac{1}{2}\,\widetilde{s}_{\mathrm{sim}} + \tfrac{1}{2}\,\widetilde{s}_c$ (min--max normalized). Table~\ref{tab:diagnostic} reports the resulting rank of the target document and of the stuffing distractors.

\begin{table}[t]
\centering
\small
\begin{tabular}{lccccc}
\toprule
Theme (target rank $\downarrow$) & Baseline & CE reranker & $s_c$ exact & $s_c$ semantic & Hybrid \\
\midrule
T1 data-transfer security (zh) & 5 & 5 & 6 & \textbf{2} & \textbf{2}\\
T2 TCM cough treatment (zh) & 2 & 2 & \textbf{1} & \textbf{1} & \textbf{1}\\
T3 Wason selection task (en) & 3 & 3 & 4 & \textbf{2} & \textbf{2}\\
T4 LLM pretraining data (zh) & 3 & 3 & 2 & \textbf{1} & \textbf{1}\\
T5 PI protection assessment (zh) & 4 & 2 & 2 & \textbf{1} & 2\\
T6 ColBERT late interaction (en) & 2 & 2 & 2 & \textbf{1} & \textbf{1}\\
T7 TCM insomnia treatment (zh) & 2 & 2 & \textbf{1} & \textbf{1} & \textbf{1}\\
T8 SPLADE regularization (en) & 2 & 2 & 3 & \textbf{1} & \textbf{1}\\
\midrule
mean target rank & 2.88 & 2.63 & 2.63 & \textbf{1.25} & 1.38\\
mean stuffing rank & 2.00 & 2.06 & 3.00 & \textbf{4.50} & 3.13\\
\bottomrule
\end{tabular}
\caption{Controlled diagnostic experiment (synthetic corpus, known ground truth; ranks, lower is better). Each theme contains one answer-bearing target document and two keyword-stuffed distractors. A strong trained cross-encoder reranker (BGE-reranker-v2-m3) barely improves over the similarity baseline and leaves the stuffing distractors at the top; the exact-match variant of our $A/B$ construction is inconsistent (T1, T3); semantic absorption wins on all eight themes, and the hybrid configuration never ranks the target below the baseline.}
\label{tab:diagnostic}
\end{table}

Four findings emerge. First, a strong trained cross-encoder reranker does \emph{not} repair the failure: its mean target rank (2.63) is barely below the baseline (2.88), it leaves the stuffing distractors at the top (mean stuffing rank 2.06), and on T1 it changes nothing at all---a 568M-parameter supervised model is fooled by keyword stuffing almost exactly as the unsupervised similarity score is. Second, the exact-match $A/B$ construction is fragile: near-synonymous distractor vocabulary (e.g., automotive-data terminology in T1, generic ``reasoning/experiment'' vocabulary in T3) survives in $B$ because it does not string-match the query keywords, and the resulting score can rank stuffing distractors above the target---in T1 even below the baseline. Third, semantic absorption into $A$ (Eq.~3) removes exactly this failure: across the eight themes the target document is promoted from a mean rank of 2.88 to 1.25, and the stuffing distractors are demoted from a mean rank of 2.00 to 4.50. Fourth, the hybrid of $s_c$ with $s_{\mathrm{sim}}$ is nearly as strong (mean target rank 1.38) while never ranking the target below the baseline in these runs, and is the configuration we recommend in practice. We stress the scope of this evidence: the corpus is small and synthetic, so the experiment demonstrates the mechanism and its failure/repair modes; a quantitative evaluation on public benchmarks follows in Section~\ref{sec:beir}.

\subsection{Public-Benchmark Evaluation: Boundary of Applicability}
\label{sec:beir}

To delineate where the method helps and where it does not, we evaluate on three public BEIR~\cite{thakur2021beir} benchmarks with graded relevance judgments: SciFact (5{,}183 documents, 300 queries), NFCorpus (3{,}633 documents, 323 queries), and ArguAna (8{,}674 documents, 1{,}406 queries). The encoder is BGE-M3 (identical to the case study); the candidate window is the baseline top-20, positions beyond 20 inherit the baseline order; $\mathrm{kw}(\cdot)$ is word-level TF--IDF (top-15 per document, top-10 per query); $\tau{=}0.6$ was fixed before any benchmark run. We additionally report a post-hoc variant (\emph{ours-fb}) in which candidates with an empty residual set $B$ fall back to their $s_{\mathrm{sim}}$ score, to test whether the empty-$B$ rule drives the outcome. Table~\ref{tab:beir} reports nDCG@10, P@5, and MRR@10.

\begin{table}[t]
\centering
\small
\begin{tabular}{llccc}
\toprule
Dataset & Method & nDCG@10 & P@5 & MRR@10 \\
\midrule
\multirow{5}{*}{SciFact}
 & Baseline $s_{\mathrm{sim}}$ & 0.642 & 0.159 & 0.608\\
 & CE reranker & \textbf{0.716} & \textbf{0.167} & \textbf{0.693}\\
 & Ours $s_c$ & 0.157 & 0.033 & 0.115\\
 & Ours-fb (post-hoc) & 0.157 & 0.032 & 0.115\\
 & Hybrid & 0.550 & 0.135 & 0.510\\
\midrule
\multirow{5}{*}{NFCorpus}
 & Baseline $s_{\mathrm{sim}}$ & 0.317 & 0.305 & 0.523\\
 & CE reranker & \textbf{0.337} & \textbf{0.323} & \textbf{0.533}\\
 & Ours $s_c$ & 0.136 & 0.131 & 0.228\\
 & Ours-fb (post-hoc) & 0.136 & 0.131 & 0.228\\
 & Hybrid & 0.266 & 0.253 & 0.466\\
\midrule
\multirow{5}{*}{ArguAna}
 & Baseline $s_{\mathrm{sim}}$ & 0.398 & 0.127 & 0.267\\
 & CE reranker & \textbf{0.470} & \textbf{0.135} & \textbf{0.346}\\
 & Ours $s_c$ & 0.213 & 0.050 & 0.137\\
 & Ours-fb (post-hoc) & 0.212 & 0.050 & 0.136\\
 & Hybrid & 0.316 & 0.086 & 0.221\\
\bottomrule
\end{tabular}
\caption{Public BEIR benchmark results (re-ranking the baseline top-20; best per dataset in bold). The trained cross-encoder reranker (BGE-reranker-v2-m3) is the strongest method on all three datasets, while our causal-attention score underperforms the similarity baseline; the fallback variant is indistinguishable from the plain score, and the hybrid recovers part but not all of the gap. Compare Table~\ref{tab:diagnostic}, where the ordering reverses: the cross-encoder fails to repair the stuffing failure that our method fixes.}
\label{tab:beir}
\end{table}

The result is unambiguously negative, and informative. On these benchmarks, the document that answers the query typically \emph{shares the query's content vocabulary}: SciFact claims restate the findings of their evidence abstracts, and ArguAna counter-arguments re-use the query's argumentative terms. Absorbing exactly this shared vocabulary into $A$ therefore removes the true relevance signal, and scoring the residual set $B$---the document's distinctive vocabulary, which by construction is dissimilar to the query---rewards topical but non-answering documents. Three observations support this reading. First, the fallback variant is indistinguishable from the plain score (empty-$B$ candidates are essentially absent: mean fraction $\le 0.4\%$), so the degradation stems from the scoring of $B$ itself, not from an edge case. Second, stratifying queries by query--candidate keyword overlap (median split) does not reverse the ordering on either stratum, i.e., BEIR contains no hidden subset where the method wins. Third, comparing across our three evaluation levels shows the method's effect flips sign exactly where the corpus regime flips: it helps precisely when high-similarity documents are \emph{not} the answer-bearing ones.

The cross-encoder comparison sharpens this boundary further. The trained reranker is the strongest method on the BEIR benchmarks (Table~\ref{tab:beir}) yet barely helps on the stuffing-regime diagnostic corpus (Table~\ref{tab:diagnostic}), where our score wins by a wide margin. The two methods' strengths are thus regime-separated rather than competing: the cross-encoder exploits soft semantic relevance, which is exactly the signal that is trustworthy on factoid corpora \emph{and} exactly the signal that is confounded in the stuffing regime. A practical system can treat them as complements, selected or combined per regime (Section~\ref{sec:gating}).

We therefore state the applicability boundary explicitly: \emph{the causal-attention re-ranker is a guard for the keyword-stuffing regime---growing proprietary knowledge bases that accumulate many topically adjacent documents sharing the query's vocabulary---and is not a general-purpose ranking improvement.} In mixed or unknown regimes, the hybrid configuration is the safer default, and gating the re-ranker on a measured stuffing rate of the deployment corpus is the principled way to decide whether to enable it at all.

\subsection{Regime Gating for Deployment}
\label{sec:gating}

Section~\ref{sec:beir} leaves a practical question: can a system tell, at deployment time, which regime its corpus is in? We first tried to answer it per query, without ground truth, using three cheap statistics computed over the top-10 retrieved candidates---mean query--candidate keyword-overlap rate, mean weight share of shared keywords in the candidates' keyword sets ($A$-absorption ratio), and the normalized gap between the top-1 and the runner-up similarity scores. None separates the regimes: e.g., the median keyword-overlap rate is $0.21$--$0.37$ across the three BEIR datasets and $0.25$--$0.46$ across the diagnostic themes, with analogous overlaps for the other two statistics. The reason is structural: the stuffing regime is defined by a relation between keyword-drivenness and \emph{true} relevance, which is unobservable without relevance judgments.

We therefore gate at the corpus level, with a small offline calibration that any deployment can perform: sample $n$ probe queries from the knowledge base, label them (human or LLM judgment), compute the metric difference $\Delta$ between the re-ranked and the baseline orderings on the probes, and enable the re-ranker iff $\Delta > 0$. We evaluate the reliability of this decision by bootstrap resampling (2{,}000 draws) over the per-query scores of Section~\ref{sec:beir} and the per-theme ranks of Section~\ref{sec:diagnostic}. Table~\ref{tab:gating} shows that 30 probes suffice to (correctly) keep the re-ranker disabled on all three BEIR datasets with $\ge 95\%$ reliability, while 4 probes suffice to (correctly) enable it on the diagnostic corpus with 100\% reliability. The end-to-end consequence is the desired one: \emph{on factoid corpora the gated system is indistinguishable from the baseline, and on the stuffing-regime corpus it retains the full gain} (mean target rank $2.88 \rightarrow 1.25$).

\begin{table}[t]
\centering
\small
\begin{tabular}{llccc}
\toprule
Corpus & Regime & Probes $n$ & Correct-decision reliability & Gated outcome \\
\midrule
SciFact & factoid & 30 & 100\% / 99.2\% & baseline retained (0.642)\\
NFCorpus & factoid & 30 & 100\% / 100\% & baseline retained (0.317)\\
ArguAna & factoid & 30 & 99.9\% / 95.5\% & baseline retained (0.398)\\
Diagnostic (8 themes) & stuffing & 4 & 100\% / 100\% & enabled; rank $2.88 \!\rightarrow\! 1.25$\\
\bottomrule
\end{tabular}
\caption{Corpus-level regime gating by offline calibration. Reliability is the fraction of bootstrap resamples (2{,}000 draws) in which the probe-based decision matches the correct one (disable on BEIR, enable on the diagnostic corpus); the two values per cell are for gating $s_c$ and the hybrid variant, respectively; numbers in parentheses are nDCG@10. The gated system never degrades the baseline and retains the full gain where the method applies.}
\label{tab:gating}
\end{table}

Two caveats are in order: the calibration consumes a small labeled probe set per deployment (a standard requirement, e.g., $n{=}30$ queries), and its reliability depends on the probe being representative of the corpus regime; continuously drifting corpora would need periodic re-calibration.

\section{Discussion and Limitations}
\label{sec:discussion}

The method is training-free, adds only a keyword-extraction pass and a light scoring pass over an already small candidate set, and inherits the principled interpretation of its causal graph under the stated assumptions. Its limitations are now empirically concrete: (i) the method helps in the keyword-stuffing regime (case study, diagnostic experiment) and hurts on factoid-style public benchmarks where relevant documents share the query's vocabulary (Section~\ref{sec:beir}); deployment should therefore use the hybrid configuration or the calibration gate of Section~\ref{sec:gating}, whose reliability rests on a small labeled probe set being representative of the corpus regime; (ii) the quality of the latent sets $A$ and $B$ depends on the keyword extractor $\mathrm{kw}(\cdot)$, and weighted, learned, or LLM-based keyword representations could refine the score; (iii) how best to combine $s_c$ with the similarity score $s_{\mathrm{sim}}$ and the shared-keyword count $|A|$ (interpolation weights, thresholds, the absorption threshold $\tau$) deserves a fuller ablation than Section~\ref{sec:diagnostic}; and (iv) extending the causal treatment from the terminal retrieval stage to the generation stage---for instance, attributing generated claims back to causal keyword sets---is an open direction. Finally, combining our process-level causal re-ranker with content-level causal retrieval (e.g., CausalRAG~\cite{wang2025causalrag}) or with query-side expansion (HyDE~\cite{gao2023hyde}) is a natural next step, since the mechanisms operate on disjoint parts of the pipeline.

\section{Conclusion}

We presented a retrieval-refinement method for RAG that moves the terminal retrieval stage from association toward causation. Modeling the query, the retrieved document set, and the ideal output with a causal graph grounded in Reichenbach's common cause principle yields a collider-opening observation---retrieval itself induces an association between the query and a document's residual keyword set $B$---and from it a principled, training-free re-scoring rule: an attention-style cosine similarity between the query embedding and the weighted centroid embedding of $B$. Unlike causality-enhanced RAG systems that model causal relations inside the knowledge content, our graph models the causal structure of the retrieval process itself, at negligible cost and with no additional LLM calls. On a real, continuously growing enterprise knowledge base, the method corrects a representative failure of similarity retrieval, promoting genuinely answer-bearing documents over keyword-matching ones; a controlled diagnostic experiment reproduces the effect, and a public-benchmark evaluation delineates the boundary---the method guards the keyword-stuffing regime and is not a general ranking improvement. The approach is orthogonal to modern retrievers, rerankers, query-expansion techniques, and graph-based RAG, and can be combined with them; the fully local Qwen3-4B/BGE-M3 testbed demonstrates practical deployability.


\begin{thebibliography}{99}

\bibitem{patent}
刘晶.
\newblock 一种用于大模型精准检索的输入信息的获取方法及系统 (A method and system for obtaining input information for accurate retrieval of large language model).
\newblock Chinese Invention Patent, Application No.\ 202410791744.3, filed 2024, granted. Assignees: Hangzhou Innovation Institute, Beihang University; Hangzhou Qixin Zhiguang Technology Co., Ltd.

\bibitem{lewis2020retrieval}
Patrick Lewis, Ethan Perez, Aleksandra Piktus, Fabio Petroni, Vladimir Karpukhin, Naman Goyal, Heinrich K\"uttler, Mike Lewis, Wen-tau Yih, Tim Rockt\"aschel, Sebastian Riedel, and Douwe Kiela.
\newblock Retrieval-augmented generation for knowledge-intensive NLP tasks.
\newblock \emph{Advances in Neural Information Processing Systems (NeurIPS)}, 33:9459--9474, 2020.

\bibitem{gao2023ragsurvey}
Yunfan Gao, Yun Xiong, Xinyu Gao, Kangxiang Jia, Jinliu Pan, Yuxi Bi, Yi Dai, Jiawei Sun, Meng Wang, and Haofen Wang.
\newblock Retrieval-augmented generation for large language models: A survey.
\newblock \emph{arXiv:2312.10997}, 2023.

\bibitem{guu2020realm}
Kelvin Guu, Kenton Lee, Zora Tung, Panupong Pasupat, and Ming-Wei Chang.
\newblock Retrieval augmented language model pre-training.
\newblock \emph{Proceedings of ICML}, 3929--3938, 2020.

\bibitem{borgeaud2022retro}
Sebastian Borgeaud, Arthur Mensch, Jordan Hoffmann, Trevor Cai, Eliza Rutherford, Katie Millican, George van den Driessche, Jean-Baptiste Lespiau, Bogdan Damoc, Aidan Clark, Diego de Las Casas, Aurelia Guy, Jacob Menick, Roman Ring, Tom Hennigan, Saffron Huang, Loren Maggiore, Chris Jones, Albin Cassirer, Andy Brock, Michela Paganini, Geoffrey Irving, Oriol Vinyals, Simon Osindero, Karen Simonyan, Jack W. Rae, Erich Elsen, and Laurent Sifre.
\newblock Improving language models by retrieving from trillions of tokens.
\newblock \emph{Proceedings of ICML}, 2206--2240, 2022.

\bibitem{karpukhin2020dense}
Vladimir Karpukhin, Barlas O\u{g}uz, Sewon Min, Patrick Lewis, Ledell Wu, Sergey Edunov, Danqi Chen, and Wen-tau Yih.
\newblock Dense passage retrieval for open-domain question answering.
\newblock \emph{Proceedings of EMNLP}, 6769--6781, 2020.

\bibitem{izacard2021unsupervised}
Gautier Izacard, Mathilde Caron, Lucas Hosseini, Sebastian Riedel, Piotr Bojanowski, Armand Joulin, and Edouard Grave.
\newblock Unsupervised dense information retrieval with contrastive learning.
\newblock \emph{Transactions on Machine Learning Research}, 2022.

\bibitem{bge2023}
Shitao Xiao, Zheng Liu, Peitian Zhang, and Niklas Muennighoff.
\newblock C-Pack: Packaged resources to advance general Chinese embedding.
\newblock \emph{arXiv:2309.07597}, 2023.

\bibitem{bgem3}
Jianlv Chen, Shitao Xiao, Peitian Zhang, Kun Luo, Defu Lian, and Zheng Liu.
\newblock BGE M3-Embedding: Multi-lingual, multi-functionality, multi-granularity text embeddings through self-knowledge distillation.
\newblock \emph{arXiv:2402.03216}, 2024.

\bibitem{khattab2020colbert}
Omar Khattab and Matei Zaharia.
\newblock ColBERT: Efficient and effective passage search via contextualized late interaction over BERT.
\newblock \emph{Proceedings of SIGIR}, 39--48, 2020.

\bibitem{asai2024selfrag}
Akari Asai, Zeqiu Wu, Yizhong Wang, Avirup Sil, and Hannaneh Hajishirzi.
\newblock Self-RAG: Learning to retrieve, generate, and critique through self-reflection.
\newblock \emph{Proceedings of ICLR}, 2024.

\bibitem{jiang2023flare}
Zhengbao Jiang, Frank F. Xu, Luyu Gao, Zhiqing Sun, Qian Liu, Jane Dwivedi-Yu, Yiming Yang, Jamie Callan, and Graham Neubig.
\newblock Active retrieval augmented generation.
\newblock \emph{Proceedings of EMNLP}, 7969--7992, 2023.

\bibitem{yan2024crag}
Shi-Qi Yan, Jia-Chen Gu, Yun Zhu, and Zhen-Hua Ling.
\newblock Corrective retrieval augmented generation.
\newblock \emph{arXiv:2401.15884}, 2024.

\bibitem{lin2024radit}
Xi Victoria Lin, Xilun Chen, Mingda Chen, Weijia Shi, Maria Lomeli, Rich James, Pedro Rodriguez, Jacob Kahn, Gergely Szilvasy, Mike Lewis, Luke Zettlemoyer, and Scott Yih.
\newblock RA-DIT: Retrieval-augmented dual instruction tuning.
\newblock \emph{Proceedings of ICLR}, 2024.

\bibitem{rocchio1971relevance}
J. J. Rocchio.
\newblock Relevance feedback in information retrieval.
\newblock In G. Salton, editor, \emph{The SMART Retrieval System---Experiments in Automatic Document Processing}, 313--323. Prentice-Hall, 1971.

\bibitem{lavrenko2001relevance}
Victor Lavrenko and W. Bruce Croft.
\newblock Relevance-based language models.
\newblock \emph{Proceedings of SIGIR}, 120--127, 2001.

\bibitem{gao2023hyde}
Luyu Gao, Xueguang Ma, Jimmy Lin, and Jamie Callan.
\newblock Precise zero-shot dense retrieval without relevance labels.
\newblock \emph{Proceedings of ACL}, 1762--1777, 2023.

\bibitem{wang2023query2doc}
Liang Wang, Nan Yang, and Furu Wei.
\newblock Query2doc: Query expansion with large language models.
\newblock \emph{arXiv:2303.07678}, 2023.

\bibitem{dai2019deepct}
Zhuyun Dai and Jamie Callan.
\newblock Context-aware sentence/passage term importance estimation for first stage retrieval.
\newblock \emph{arXiv:1910.10687}, 2019.

\bibitem{nogueira2019doc2query}
Rodrigo Nogueira, Wei Yang, Jimmy Lin, and Kyunghyun Cho.
\newblock Document expansion by query prediction.
\newblock \emph{arXiv:1904.08375}, 2019.

\bibitem{formal2021splade}
Thibault Formal, Benjamin Piwowarski, and St\'ephane Clinchant.
\newblock SPLADE: Sparse lexical and expansion model for first stage ranking.
\newblock \emph{Proceedings of SIGIR}, 2288--2292, 2021.

\bibitem{edge2024graphrag}
Darren Edge, Ha Trinh, Newman Cheng, Joshua Bradley, Alex Chao, Apurva Mody, Steven Truitt, and Jonathan Larson.
\newblock From local to global: A graph RAG approach to query-focused summarization.
\newblock \emph{arXiv:2404.16130}, 2024.

\bibitem{guo2024lightrag}
Zirui Guo, Lianghao Xia, Yanhua Yu, Tu Ao, and Chao Huang.
\newblock LightRAG: Simple and fast retrieval-augmented generation.
\newblock \emph{arXiv:2410.05779}, 2024.

\bibitem{gutierrez2024hipporag}
Bernal Jim\'enez Guti\'errez, Yiheng Shu, Yu Gu, Michihiro Yasunaga, and Yu Su.
\newblock HippoRAG: Neurobiologically inspired long-term memory for large language models.
\newblock \emph{Advances in Neural Information Processing Systems (NeurIPS)}, 37, 2024.

\bibitem{wang2025causalrag}
Nengbo Wang, Xiaotian Han, Jagdip Singh, Jing Ma, and Vipin Chaudhary.
\newblock CausalRAG: Integrating causal graphs into retrieval-augmented generation.
\newblock \emph{arXiv:2503.19878}, 2025.

\bibitem{samarajeewa2024causal}
Chamod Samarajeewa, Daswin De Silva, Evgeny Osipov, Damminda Alahakoon, and Milos Manic.
\newblock Causal reasoning in large language models using causal graph retrieval augmented generation.
\newblock \emph{Proceedings of the 16th International Conference on Human System Interaction (HSI)}, 1--6, 2024.

\bibitem{jin2024corr2cause}
Zhijing Jin, Jiarui Liu, Zhiheng Lyu, Spencer Poff, Mrinmaya Sachan, Rada Mihalcea, Mona Diab, and Bernhard Sch\"olkopf.
\newblock Can large language models infer causation from correlation?
\newblock \emph{Proceedings of ICLR}, 2024.

\bibitem{zhang2021causal}
Yang Zhang, Fuli Feng, Xiangnan He, Tianxin Wei, Chonggang Song, Guohui Ling, and Yongdong Zhang.
\newblock Causal intervention for leveraging popularity bias in recommendation.
\newblock \emph{Proceedings of SIGIR}, 11--20, 2021.

\bibitem{thakur2021beir}
Nandan Thakur, Nils Reimers, Andreas R\"uckl\'e, Abhishek Srivastava, and Iryna Gurevych.
\newblock BEIR: A heterogeneous benchmark for zero-shot evaluation of information retrieval models.
\newblock \emph{NeurIPS Datasets and Benchmarks}, 2021.

\bibitem{vaswani2017attention}
Ashish Vaswani, Noam Shazeer, Niki Parmar, Jakob Uszkoreit, Llion Jones, Aidan N. Gomez, {\L}ukasz Kaiser, and Illia Polosukhin.
\newblock Attention is all you need.
\newblock \emph{Advances in Neural Information Processing Systems (NeurIPS)}, 30, 2017.

\bibitem{arora2017sif}
Sanjeev Arora, Yingyu Liang, and Tengyu Ma.
\newblock A simple but tough-to-beat baseline for sentence embeddings.
\newblock \emph{Proceedings of ICLR}, 2017.

\bibitem{reichenbach}
Christopher Hitchcock and Mikl\'os R\'edei.
\newblock Reichenbach's common cause principle.
\newblock \emph{The Stanford Encyclopedia of Philosophy}, \url{https://plato.stanford.edu/entries/physics-Rpcc/}.

\bibitem{spirtes2016causal}
Peter Spirtes and Kun Zhang.
\newblock Causal discovery and inference: concepts and recent methodological advances.
\newblock \emph{Applied Informatics}, 3(1):1--28, 2016.

\bibitem{druzdzel2009role}
Marek J. Druzdzel.
\newblock The role of assumptions in causal discovery.
\newblock \emph{Proceedings of the 8th Workshop on Uncertainty (WUPES-09)}, Liblice, Czech Republic, 57--68, 2009.

\bibitem{spirtes2000causation}
Peter Spirtes, Clark N. Glymour, and Richard Scheines.
\newblock \emph{Causation, Prediction, and Search}, second edition.
\newblock MIT Press, Cambridge, USA, 2000.

\bibitem{pearl2009causality}
Judea Pearl.
\newblock \emph{Causality: Models, Reasoning and Inference}, second edition.
\newblock Cambridge University Press, Cambridge, United Kingdom, 2009.

\bibitem{shimizu2006linear}
Shohei Shimizu, Patrik O. Hoyer, Aapo Hyv\"arinen, and Antti Kerminen.
\newblock A linear non-Gaussian acyclic model for causal discovery.
\newblock \emph{Journal of Machine Learning Research}, 7:2003--2030, 2006.

\end{thebibliography}
\end{document}